\pdfoutput=1

\documentclass[11pt]{article}

\usepackage[final]{acl}

\usepackage{times}
\usepackage{latexsym}

\usepackage[T1]{fontenc}

\usepackage[utf8]{inputenc}
\usepackage{hyperref}       
\usepackage{url}            
\usepackage{booktabs}       
\usepackage{amsfonts}       
\usepackage{nicefrac}       
\usepackage{microtype}      
\usepackage{xcolor}         
\usepackage{amsmath}
\usepackage{amssymb}
\usepackage{multirow}
\usepackage[ruled]{algorithm2e}
\usepackage{float}
\usepackage{ulem}
\usepackage{microtype}
\usepackage{inconsolata}
\usepackage{graphicx}
\usepackage{subfigure}
\usepackage{makecell}
\usepackage{bbding}
\definecolor{color1}{HTML}{F2CFEE}
\definecolor{color2}{HTML}{DCEAF7}
\definecolor{QQQ}{HTML}{FFBE7A}
\definecolor{III}{HTML}{D76364}
\definecolor{RRR}{HTML}{54B345}
\definecolor{PPP}{HTML}{B883D4}
\definecolor{AAA}{HTML}{82B0D2}
\usepackage{pifont}

\title{WarriorCoder: Learning from Expert Battles to Augment Code Large Language Models}

\author{Huawen Feng$^{1,2}$, Pu Zhao$^{2}$, Qingfeng Sun$^{2}$, Can Xu$^{2}$, Fangkai Yang$^{2}$, \\ 
\textbf{Lu Wang$^{2}$, Qianli Ma$^{1}$, Qingwei Lin$^{2}$, Saravan Rajmohan$^{2}$, Dongmei Zhang$^{2}$, Qi Zhang$^{2}$}  \\
  $^{1}$School of Computer Science and Engineering, South China University of Technology, China\\
  $^{2}$Microsoft\\
  \texttt{qianlima@scut.edu.cn}\\
  \texttt{\{v-huawenfeng,puzhao,qins,caxu,fangkai.yang,wlu,qlin,saravan.rajmohan,}\\
  \texttt{dongmeiz,zhang.qi\}@microsoft.com}
}

\begin{document}
\maketitle
\begin{abstract}
Despite recent progress achieved by code large language models (LLMs), their remarkable abilities are largely dependent on fine-tuning on the high-quality data, posing challenges for data collection and annotation. To address this, current methods often design various data flywheels to collect complex code instructions, enabling models to handle more intricate tasks. However, these approaches typically rely on off-the-shelf datasets and data augmentation from a limited set of proprietary LLMs (e.g., Claude, GPT4, and so on), which restricts the diversity of the constructed data and makes it prone to systemic biases. In this paper, we propose \textbf{WarriorCoder}, a novel paradigm learns from expert battles to address these limitations. Specifically, we create an arena where leading expert code LLMs challenge each other, with evaluations conducted by impartial judges. This competitive framework generates novel training data from scratch, leveraging the strengths of all participants. Experimental results show that \textbf{WarriorCoder} achieves state-of-the-art performance compared to previous models of the same size, even without relying on proprietary LLMs.
\end{abstract}

\section{Introduction}
\label{sec:Introduction}
Recent large language models (LLMs) have demonstrated impressive performance on code-related tasks~\cite{StarCoder, CodeLlama, DeepSeek-Coder, DeepSeek-Coder-V2, AlphaCode, CodeGen, CodeGeeX, InCoder, CodeT5}. These successes highlight that pre-training on vast amounts of code data significantly enhances their core coding abilities. In addition to pre-training, several approaches that fine-tune LLMs with instruction-following data~\cite{Enhancing} have also made substantial progress in improving models' understanding of user instructions and the quality of their responses. However, the effectiveness of post-training is heavily dependent on the availability of high-quality data~\cite{WizardLM}, and challenges of data collection and annotation remain difficult to overcome.

\begin{figure}[t!]
	\centering
	\includegraphics[width=1.0\linewidth]{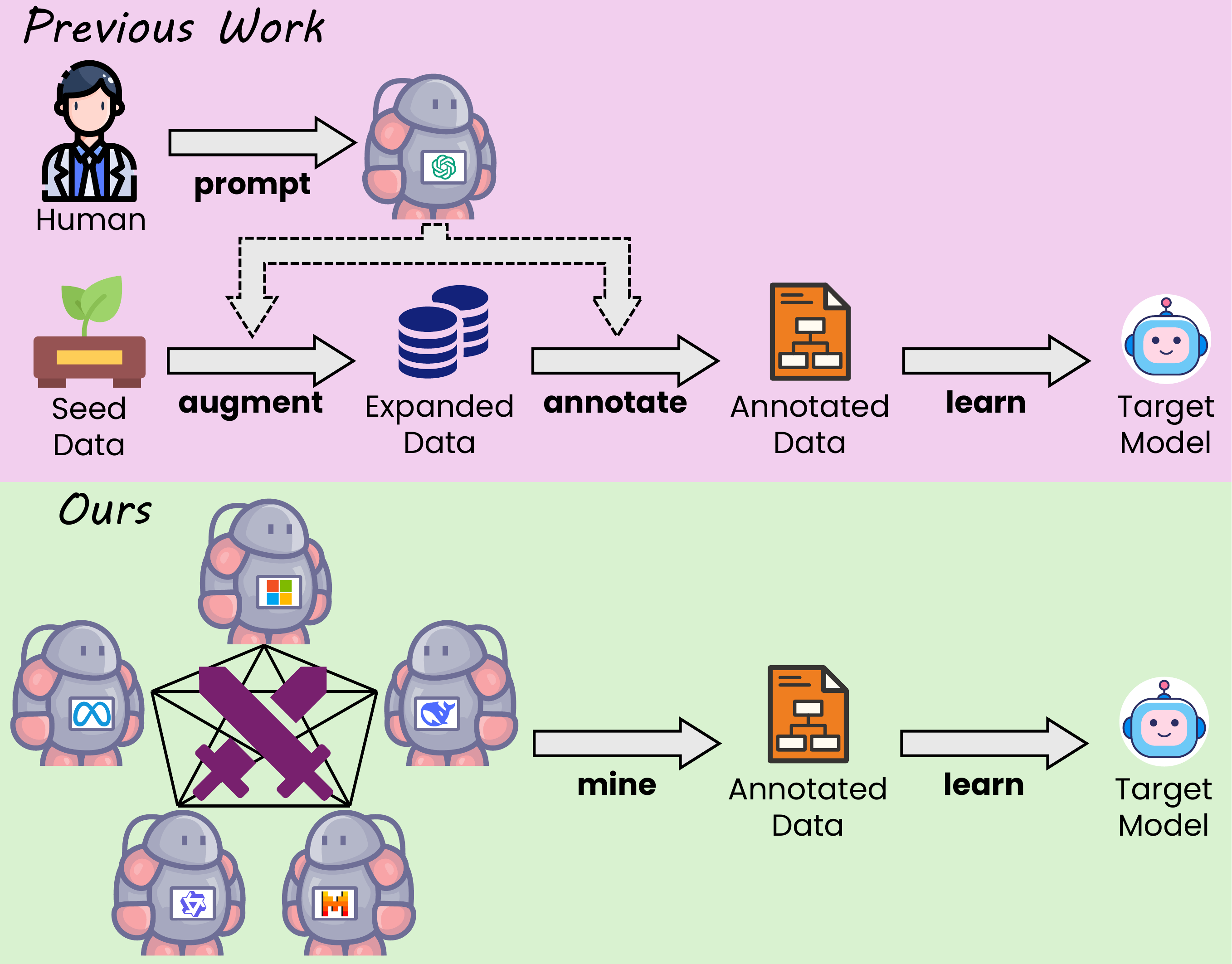}
    \caption{The comparisons between our method and traditional data flywheels. Unlike previous work, we guides the target model to learn from pairwise competitions. No demand for seed datasets, human-generated prompts, or annotations from proprietary models, the target model integrates the strengths of its competitors.}
    \label{Fig:intro}
\end{figure}

To address these challenges, some approaches propose various data flywheels to generate instruction data. Building on Self-Instruct, \citet{codealpaca} constructs Code Alpaca by prompting teacher LLMs to generate instructions in a few-shot setting. To further enhance the diversity and complexity of Code Alpaca, WizardCoder~\cite{WizardCoder} employs Evol-Instruct to evolve the original instructions. These methods apply general data augmentation to instruction construction, lacking specific design considerations for the code domain. Given that, recent methods specifically design frameworks for instruction generation tailored to code. For example, WaveCoder~\cite{WaveCoder} collects raw code snippets and defines different tasks based on them. Similarly, InverseCoder~\cite{InverseCoder} prompts code LLMs to generate high-quality instructions based on the original code through techniques like code summarization and self-evaluation. These methods construct data for code in various ways, effectively enhancing the model’s code generation capabilities. However, they still rely on existing datasets~\cite{OctoPack} and calls for proprietary LLMs (e.g., GPT-3.5, GPT-4, etc.), making data collection costly. Additionally, the limited data sources and annotators constrain the diversity~\cite{Diversity, Self-Instruct} of the data and inherit the system biases inherent in the limited pool of annotators~\cite{Magicoder}.

The challenges mentioned above motivated us to propose \textbf{WarriorCoder}, which learns from expert battles to overcome current limitations. As illustrated in Figure~\ref{Fig:intro}, the attacker challenges the opponent within its area of expertise, and the target model learns from the winner of these pairwise competitions. Specifically, we design a completion-based method to mine the capabilities which the attacker has already mastered, then integrate Elo Rating and voting results to balance the local and global evaluation. This approach enables us to generate novel training data from scratch, incorporating the strengths of all the expert code LLMs, rather than relying on limited proprietary LLMs to expand existing datasets. Moreover, our method eliminates the need for human involvement and proprietary LLMs in the data collection, making it possible to collect high-quality, diverse data at a low cost. The main contributions of this paper are summarized as follows:
\begin{itemize}
\item We identify the limitations of current data flywheel and propose a new scalable paradigm where the target model learns from expert battles to solve them.
\item We design a completion-based method for collecting instructions and introduce the Elo Rating system for evaluating responses, enabling the creation of high-quality and diverse training data at a low cost. Fine-tuned on this data, \textbf{WarriorCoder} incorporates the strengths of all the experts, achieving state-of-the-art performance compared to previous models of the same size, without relying on proprietary LLMs.
\item Extensive experiments demonstrate the excellent performance of \textbf{WarriorCoder} on multiple code-related tasks, with ablation and analysis studies explaining how and why it works.
\end{itemize}

\section{Related Work}

\subsection{Code LLMs}
Code plays a crucial role in application areas for LLMs~\cite{LiveCodeBench}, attracting significant interest from both academia and industry. Codex~\cite{codex}, an LLM with 12 billion parameters, can solve 72.31\% of complex Python programming problems. Following the success of Codex~\cite{Beyond}, the rise of new code LLMs has demonstrated even greater capabilities, such as code generation and debugging, as model sizes continue to grow~\cite{SoftwareEngineering, NL2Code}. Despite this impressive progress, the performance of current open-source models still lags behind that of proprietary ones (e.g., GPT-3.5, GPT-4, etc.), primarily because stronger models often keep their training data proprietary~\cite{Qwen2.5}. As a result, the lack of publicly available code datasets remains a significant barrier to further development in this field.

\subsection{Learning from Battles} 
Studying how people interact with LLMs in real-world scenarios is a pressing need for ensuring the alignment of LLMs~\cite{Platform}. The LMSYS Chatbot Arena~\cite{LMSYS-Chat-1M} has emerged as a groundbreaking initiative for exploring real-world LLM-user interactions, collecting and analyzing data from an open platform with over 240K votes. Experimental results demonstrate that the quality of data from the LMSYS Chatbot Arena is competitive with that of ShareGPT~\cite{vicuna}, underscoring its value for training. Now more and more attentions are paid on learning from the battles between LLMs~\cite{Arena-Hard, Open-LLM-Leaderboard, LongCodeArena}. However, collecting data through human online evaluations is both expensive and time-consuming. To address this, recent work has leveraged LLMs to provide their preferences when faced with different responses~\cite{ArenaLearning, AutoArena}. Although these methods eliminate the need for human annotation during data collection, they still require pre-designed, high-quality instructions.

\subsection{LLM as a Judge}
Offering an automatic alternative to the scalability challenges inherent in human evaluation, the concept of LLM-as-a-judge has garnered significant public attention in recent years~\cite{CanAlternative}. As large language models (LLMs) such as GPT-4 have demonstrated impressive capabilities, they are increasingly being considered for use in evaluating other machine-generated outputs~\cite{CodeUltraFeedback}. Experimental results by \citet{JudgingJudges} show that these strong LLMs are capable of achieving a Cohen's Kappa coefficient over 80\%, a metric typically used to assess the level of agreement between human raters. This performance level is comparable to the consensus found among human experts, highlighting the potential of LLMs to serve as reliable evaluators in various contexts. However, judge models often struggle with complex problems, and evaluating responses to such problems can be just as challenging as answering them~\cite{Meta-Rewarding}. Moreover, LLM-as-a-judge can introduce system biases, such as position bias, verbosity bias, and self-enhancement bias, which can undermine the fairness of the evaluation process~\cite{JudgeBias, JudgingLLM-as-a-Judge}.

\section{WarriorCoder: Learning from Expert Battles}
\begin{figure*}[t!]
	\centering
	\includegraphics[width=1.0\linewidth]{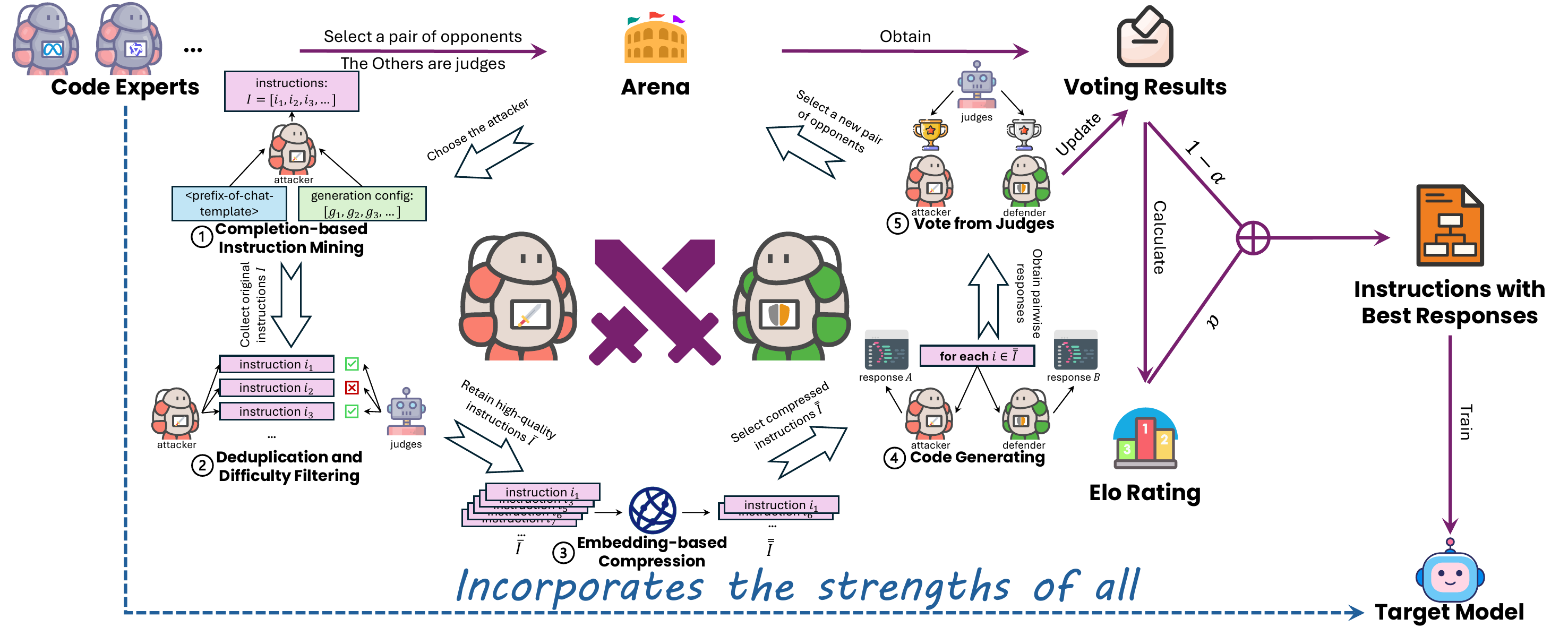}
    \caption{The diagram of learning from expert battles. In each round of the arena, the attacker challenges the defender in its area of expertise under the evaluation of judges, and then the winner's response is added to the training data. In this manner, the target model gradually incorporates the strengths of all the code experts by fine-tuning on the data.}
    \label{Fig:method}
\end{figure*}

In this section, we describe how \textbf{WarriorCoder} learns from expert battles. Unlike previous approaches that expand existing datasets by prompting a limited pool of proprietary LLMs, we construct an arena where state-of-the-art code LLMs compete against each other. Each model leverages its learned knowledge to challenge others, while judges evaluate the outcomes. The target model then learns from the winner of these pairwise competitions, progressively integrating the strengths of all competitors.

\subsection{Competitors Setting}
The capabilities of competitors determine the final performance of \textbf{WarriorCoder}. Theoretically, the more diverse and high-quality training data are derived from a larger and stronger pool of competitors. For this study, we select five leading open-source code experts from the BigCodeBench Leaderboard~\cite{BigCodeBench} - Athene-V2-Chat~\cite{su2025tokenassortedmixinglatent}, DeepSeek-Coder-V2-Lite-Instruct~\cite{DeepSeek-Coder-V2}, Llama-3.3-70B-Instruct~\cite{llama3herd}, Qwen2.5-72B-Instruct~\cite{Qwen2.5}, and QwQ-32B-Preview~\cite{qwq-32b-preview}. Notably, while \textbf{WarriorCoder} achieves state-of-the-art performance based solely on open-source code LLMs, it can also learn from powerful proprietary LLMs. In each round of the arena, only one pair of code experts is selected as competitors, while the remaining ones serve as judges.

\subsection{Instruction Mining from Scratch}
\label{sec:instruction}
Considering a battle between LLM A and LLM B where A is the attacker and B is the defender. The first step of the arena is to use the strengths of A to challenge B, which makes it necessary to know what A has learned during its training process. However, almost all open-source LLMs keep their core data proprietary. Inspired by Magpie~\cite{magpie}, we design a completion-based method to mine the capabilities which the code LLMs have already mastered (\ding{172} \textbf{Completion-based Instruction Mining}). Here we take Qwen2.5~\cite{Qwen2.5} as an example. A conversation about writing Python code in the chat template of Qwen2.5 is:
\begin{center}
\resizebox{0.9\linewidth}{!}{
\fcolorbox{black}{color2}{\parbox{1.0\linewidth}{<|im\_start|>system\\
You are an AI assistant designed to provide helpful on Python coding problems.<|im\_end|>\\
\textcolor{RRR}{<|im\_start|>user}\\
\textcolor{RRR}{Write a Quicksort algorithm.<|im\_end|>}\\
\textcolor{red}{<|im\_start|>assistant}\\
\textcolor{red}{Here is the solution:}\\
\textcolor{red}{def quicksort(arr):}\\
\textcolor{red}{\quad\quad...}\\
\textcolor{red}{<|im\_end|>}
}
}
}
\end{center}

Such chat templates are pre-defined conversational structures to guide the interaction between the model and the user. Based on the LLMs' strong completion abilities, we feed only the prefix of the chat template into them, prompting the LLM to generate the user instructions:

\begin{center}
\resizebox{0.9\linewidth}{!}{
\fcolorbox{black}{color2}{\parbox{1.0\linewidth}{<|im\_start|>system\\
You are an AI assistant designed to provide responses on Python coding problems.<|im\_end|>\\
\textcolor{RRR}{<|im\_start|>user}
}
}
}
\end{center}

In this way, we can collect various instructions $I$ the model has already learned under various generation settings (different values of temperature and top-p). Unlike traditional data synthesis, $I$ is not synthesized by the models but directly sampled from their distributions, which avoids pattern overfitting and significant shifts in the output distribution~\cite{Unveiling}. However, these instructions may be repetitive, ambiguous, unclear, or too easy. To address these concerns, we deduplicate the data and adopt judges to assess their difficulty (\ding{173} \textbf{Deduplication and Difficulty Filtering}). We divide the difficulty of instructions into four levels:
\begin{itemize}
\item \textbf{Excellent (9-10):} For instructions that are very clear, specific, and well-articulated. These instructions are particularly challenging and excellently designed to assess the AI's proficiency.
\item \textbf{Good (6-8):} For instructions that are clear and specific instructions. These are not overly difficult to answer and moderately assess the AI's capabilities.
\item \textbf{Average (3-5):} For instructions that are fairly clear and specific instructions. These instructions are easy to answer.
\item \textbf{Poor (1-2):} For instructions that are ambiguous or unclear.
\end{itemize}
Only good and excellent instructions are considered during the following steps:
\begin{equation}
\begin{aligned}
\bar{I} = \{i|i \in I \wedge d(i)\ge6 \}
\end{aligned}
\label{diff}
\end{equation}
where $d(i)$ is the difficulty of instruction $i$.

Then we compress the high-quality instructions $\bar{I}$ for the efficiency of post-training (\ding{174} \textbf{Embedding-based Compression}). To ensure the diversity and representativeness of instructions, we employ KCenterGreedy algorithm~\cite{kcenter} to select the final instructions $\bar{\bar{I}}$ based on the embedding model - \textit{all-roberta-large-v1}~\cite{RoBERTa}. 

\subsection{Win-Loss Decision}
The defender is required to respond to the attacker’s question, while the attacker A must also provide an answer to its own instruction (\ding{175} \textbf{Code Generating}). Once both answers are collected, the judges (the rest LLMs in arena) will evaluate the correctness and helpfulness of the pairwise responses and vote for their preferred one (more details can be found in Appendix~\ref{sec:fairness}). Then we can calculate the \textit{local score} for each response:
\begin{equation}
\begin{aligned}
x^i_{A>B} = \frac{t_{A}}{t_{A}+t_{B}} \quad x^i_{B>A} = \frac{t_{B}}{t_{A}+t_{B}}
\end{aligned}
\label{vote}
\end{equation}
where $x^i_{A>B}$ and $x^i_{B>A}$ are the local scores for A's and B's responses to the instruction $i$. $x^i_{A>B}$ represents the percentage of votes that candidate A receives, while $x^i_{B>A}$ similarly represents the percentage of votes that candidate B receives. $t_{A}$ and $t_{B}$ are the number of votes which A and B win.

However, relying solely on the \textit{local score} to select the winner can be problematic. In some cases, a weaker model may receive more votes than a stronger one, even though its responses are not significantly better. This can occur because the \textit{local score} may not fully capture the quality of the model’s performance, especially in situations where the voting is influenced by factors, such as randomness or bias from LLM judges.

To address this limitation, we propose considering both local contingency and global consistency in the decision-making process. Instead of directly basing our analysis on the immediate voting outcomes, we introduce the concept of the \textit{global score} — specifically, the Elo rating~\cite{harmless}, which provides a more comprehensive reflection of a model's relative performance over time and across various evaluations. The Elo rating system, originally developed to calculate the relative skill levels of players in two-player games (such as chess), has been successfully adapted to assess the performance of competitors in a range of competitive scenarios, including esports and other skill-based games.

By incorporating the Elo rating, we account for both local performance in individual contests and global performance across multiple rounds, providing a more robust and accurate measure of a model's overall ability. This helps to mitigate the risk of weak models winning based on isolated, potentially unrepresentative votes:
\begin{equation}
\begin{aligned}
&X^{Elo}_{A>B} = \frac{1}{1+{10}^{(R_B-R_A)/400}} \\
&X^{Elo}_{B>A} = \frac{1}{1+{10}^{(R_A-R_B)/400}} \\
\end{aligned}
\label{elo}
\end{equation}
where $X^{Elo}_{A>B}$ and $X^{Elo}_{B>A}$ indicate the expected probabilities of A defeating B and B defeating A, respectively. $R_A$ and $R_B$ are the Elo rating of A and B, which are updated dynamically and iteratively. Given the battle result of A and B on an instruction $i$, we update them by:
\begin{equation}
\begin{aligned}
R_A \gets R_A + K \times (s^i_{A>B}-X^{Elo}_{A>B}) \\
R_B \gets R_B + K \times (s^i_{B>A}-X^{Elo}_{B>A})
\end{aligned}
\label{update}
\end{equation}
where $s^i_{A>B}$ and $s^i_{B>A}$ are the actual score of the battle result of player A and B (1 for a win, 0.5 for a draw, and 0 for a loss). The factor $K$ controls the sensitivity of rating changes.

Based on Equation~\ref{vote} and Equation~\ref{elo}, we can obtain the final score of A's response for instruction $i$:
\begin{equation}
\begin{aligned}
e^{i}_{A} = \sum_{B \in Com \setminus A} \alpha X^{Elo}_{A>B} + (1-\alpha) x^i_{A>B}
\end{aligned}
\label{final}
\end{equation}

where $Com$ is the set of all the competitors and `$\setminus$' is the subtraction operation. $\alpha$ is the coefficient to balance the local contingency and global consistency.


\subsection{Final Training}
Each item in the constructed dataset consists of an instruction, responses from various strong LLMs, and their corresponding scores, which supports multiple post-training methods. We select the response with the highest score as the gold output to obtain \textbf{Instructions with Best Responses} in Figure~\ref{Fig:method} and use SFT to train the target model. In this manner, \textbf{WarriorCoder} integrates the strengths of all the code experts, as their expertise is embedded in the instructions and responses within the training data.

\section{Experiments}

\begin{table*}[t!]
    \centering
    \resizebox{0.8\linewidth}{!}{
    \begin{tabular}{ccccccc} \hline
         \multicolumn{2}{c}{Models}& {HumanEval}&  {HumanEval+}& {MBPP}& {MBPP+}& {\makecell[c]{Rely on proprietary\\LLMs?}}\\
         \hline
         \multirow{4}{*}{Proprietary Models}& {Code-Davinci-002}& \textcolor{gray}{47.0}&  \textcolor{gray}{-}&  \textcolor{gray}{58.1}& \textcolor{gray}{-}& {-}\\
         {}& {Code-Cushman-001}& \textcolor{gray}{33.5}&  \textcolor{gray}{-}&   \textcolor{gray}{45.9}& \textcolor{gray}{-}& {-}\\
         {}& {GPT-3.5-Turbo}& \textcolor{gray}{76.8}&  \textcolor{gray}{70.7}&   \textcolor{gray}{82.5}& \textcolor{gray}{69.7}& {-}\\
         {}& {GPT-4-Turbo}& \textcolor{gray}{90.2}&  \textcolor{gray}{86.6}&   \textcolor{gray}{85.7}& \textcolor{gray}{73.3}& {-}\\
         \hline
         \multirow{3}{*}{\makecell[c]{Base Models}}& {DeepSeekCoder-Base (6.7B)}& {47.6}&  {39.6}&  {70.2}& {56.6}& {-}\\
         {}& {CodeLlama (6.7B)}& {37.8}&  {35.4}&  {59.5}& {46.8}& {-}\\
         {}& {StarCoder (15B)}& {34.1}&  {29.3}&  {55.1}& {46.1}& {-}\\
         \hline 
         \multirow{10}{*}{\makecell[c]{Fine-tuned Models}}& {CodeT5+ (16B)}& {31.7}&  {26.2}&   {54.6}& {44.4}& \textcolor{green}{\Checkmark}\\
         {}& {WizardCoder-CL (6.7B)}& {48.2}&  {40.9}&   {56.6}& {47.1}& \textcolor{green}{\Checkmark}\\
         {}& {WizardCoder-SC (15B)}& {51.9}&  {45.1}&   {61.9}& {50.6}& \textcolor{green}{\Checkmark}\\
         {}& {Magicoder-DS (6.7B)}& {66.5}&  {60.4}&   {75.4}& {61.9}& \textcolor{green}{\Checkmark}\\
         {}& {MagicoderS-DS (6.7B)}& {76.8}&  {70.7}&   {75.7}& {64.4}& \textcolor{green}{\Checkmark}\\
         {}& {Magicoder-CL (6.7B)}& {60.4}&  {55.5}&   {64.2}& {52.6}& \textcolor{green}{\Checkmark}\\
         {}& {MagicoderS-CL (6.7B)}& {70.7}&  {66.5}&   {68.4}& {56.6}& \textcolor{green}{\Checkmark}\\
         {}& {WaveCoder-DS (6.7B)}& {72.0}&  {-}&  {63.6}& {-}& \textcolor{green}{\Checkmark}\\
         {}& {WaveCoder-CL (6.7B)}& {48.1}&  {-}&  {47.2}& {-}& \textcolor{green}{\Checkmark}\\
         {}& {WaveCoder-SC (15B)}& {50.5}&  {-}&  {51.0}& {-}& \textcolor{green}{\Checkmark}\\
         \hline
         {Ours}&  {WarriorCoder (6.7B)}& {\textbf{80.5} (+32.9)}&  {\textbf{75.6} (+36.0)}&  {\textbf{76.2} (+6.0)}& {\textbf{64.8} (+8.2)}& \textcolor{red}{\XSolidBrush}\\
         \hline
    \end{tabular}
    }
    \caption{The pass@1(\%) results on the code generation benchmarks (Humaneval, Humaneval+, MBPP and MBPP+).}
    \label{tab:overall}
\end{table*}

\subsection{Experimental Details}

\paragraph{Backbones} We use DeepSeekCoder-Base-6.7B~\cite{DeepSeek-Coder} to initialize \textbf{WarriorCoder}. As for the competitors of expert battles, we choose strong open-source LLMs including Athene-V2-Chat~\cite{su2025tokenassortedmixinglatent}, DeepSeek-Coder-V2-Lite-Instruct~\cite{DeepSeek-Coder-V2}, Llama-3.3-70B-Instruct~\cite{llama3herd}, Qwen2.5-72B-Instruct~\cite{Qwen2.5}, and QwQ-32B-Preview~\cite{qwq-32b-preview}.

\paragraph{Datasets}
To evaluate the code generation capability of \textbf{WarriorCoder}, we conduct evaluations on HumanEval~\cite{codex}, MBPP~\cite{mbpp}, HumanEval+~\cite{evalplus}, and MBPP+~\cite{evalperf}. Besides, we also evaluate its code reasoning and libraries usage capabilities on CRUXEval~\cite{cruxeval} and DS-1000~\cite{DS1000}. For a fair comparison, we use the same decoding strategies and generation configs as the previous work~\cite{Magicoder, WizardCoder, WaveCoder}.

\paragraph{Experimental Settings} During the intruction minging, we adopt 9 different generation configs where temperature $t \in \{1.0,1.1,1.2\}$ and top-p $p \in \{0.99,0.995,1.0\}$. The final number of battle rounds is set to 70,000 and $K$ is set to 40. $\alpha$ is set to 0.7 because we need the Elo Rating only when judges' opinions are divided on the evaluation. The detailed prompts can be found in Appendix~\ref{sec:prompt}. As for the training stage, we conduct parallel training on 8 NVIDIA A800 80G GPUs. The global batch size is set to 512, and the number of total training steps is set to 448. We use a learning rate of \(1 \times 10^{-5}\) and a weight decay of \(3 \times 10^{-7}\). Additionally, a WarmupLR scheduler with a warmup ratio of 0.2 is used.

\paragraph{Baselines} The baselines consist of proprietary models, base models, and fine-tuned models.

\textit{Proprietary Models.} These models, unlike open-source models, are developed, owned, and managed by a private entity or organization. They are trained on specialized or private datasets that are not publicly available to serve specific business needs or objectives. Access to these models is usually based on API calls. Proprietary Models include Code-Davinci-002, Code-Cushman-001, GPT-3.5-Turbo~\cite{chatgpt} and GPT-4-Turbo~\cite{gpt4}.

\textit{Base Models.} They are the foundational, pre-trained models that serve as the core for further fine-tuning or adaptation to code tasks. Base Models include DeepSeekCoder-Base~\cite{DeepSeek-Coder}, CodeLlama~\cite{CodeLlama}, and StarCoder~\cite{StarCoder}.

\textit{Fine-tuned Models.} These models are initially pre-trained on a large, general-purpose dataset and then fine-tuned on a smaller, code-specific dataset. This two-step process enhances the model's performance on coding tasks by enabling it to leverage both broad general knowledge and more focused, domain-specific expertise. Fine-tuned models include CodeT5+, DeepSeek-Coder-Instruct~\cite{DeepSeek-Coder}, WizardCoder~\cite{WizardCoder}, Magicoder~\cite{Magicoder}, and WaveCoder~\cite{WaveCoder}. The suffixes -DS, -CL, and -SC denote the base models DeepSeekCoder-Base, CodeLlama-Python, and StarCoder, respectively.

\subsection{Main Results}

\begin{table*}[t!]
    \centering
    \resizebox{0.7\linewidth}{!}{
    \begin{tabular}{cccccc} \hline
         \multicolumn{2}{c}{\multirow{2}{*}{Models}}& \multicolumn{2}{c}{CRUXEval-I}&   \multicolumn{2}{c}{CRUXEval-O}\\
         {}& {}& {Pass@1}&  {Pass@5}& {Pass@1}& {Pass@5}\\
         \hline
         \multirow{3}{*}{Proprietary Models}& {GPT-4-Turbo}& \textcolor{gray}{69.8}&  \textcolor{gray}{76.8}&  \textcolor{gray}{68.7}& \textcolor{gray}{73.0}\\
         {}& {GPT-3.5-Turbo}& \textcolor{gray}{49.0}&  \textcolor{gray}{63.2}&   \textcolor{gray}{49.4}& \textcolor{gray}{59.3}\\
         {}& {Claude-3-Opus}& \textcolor{gray}{64.2}&  \textcolor{gray}{-}&   \textcolor{gray}{65.8}& \textcolor{gray}{-}\\
         \hline
         \multirow{10}{*}{\makecell[c]{Open-source Models}}& {StarCoder (6.7B)}& {29.7}&  {47.3}&   {32.2}& {44.9}\\
         {}& {StarCoder (15B)}& {31.3}&  {49.2}&   {34.2}& {47.1}\\
         {}& {DeepSeekCoder-Instruct (6.7B)}& {37.4}&  {53.3}&   {41.2}& {52.8}\\
         {}& {CodeLlama-Python (6.7B)}& {37.3}&  {57.0}&   {35.9}& {48.8}\\
         {}& {CodeLlama-Python (13B)}& {39.7}&  {56.9}&   {39.8}& {52.5}\\	 	
         {}& {CodeLlama-Python (34B)}& {\textbf{43.9}}&  {59.5}&   {41.4}& {52.9}\\	 	
         {}& {Mistral (6.7B)}& {35.0}&  {52.3}&   {34.3}& {48.6}\\	 	
         {}& {WizardCoder (13B)}& {36.5}&  {51.6}&   {41.3}& {52.4}\\
         {}& {WizardCoder (34B)}& {42.7}&  {57.5}&   {43.4}& {53.8}\\	 		
         {}& {Magicoder(6.7B)}& {41.7}&  {62.4}&  {44.4}& {57.5}\\
         \hline
         {Ours}& {WarriorCoder (6.7B)}& {42.9}&  {\textbf{66.5}}&  {\textbf{45.4}}& {\textbf{66.3}}\\
         \hline
    \end{tabular}
    }
    \caption{The pass@1(\%) and pass@5(\%) results on the code reasoning benchmark (CRUXEval).}
    \label{tab:crux}
\end{table*}

\begin{table*}[t!]
    \centering
    \resizebox{0.75\linewidth}{!}{
    \begin{tabular}{c|cccccccc} \hline
         \multirow{2}{*}{Models}& {Matplotlib}& {NumPy}& {Pandas}& {PyTorch}& {SciPy}& {Sklearn}& {TensorFlow}& {Overall}\\
         {}& {(155)}& {(220)}& {(291)}& {(68)}& {(106)}& {(115)}& {(45)}& {(1000)}\\
         \hline
         {INCODER (6.7B)}& {28.3}& {4.4}& {3.1}& {4.4}& {2.8}& {2.8}& {3.8}& {7.4}\\
         {CodeGen-Mono (16B)}& {31.7}& {10.9}& {3.4}& {7.0}& {9.0}& {10.8}& {15.2}& {11.7}\\
         {Code-Cushman-001}& {40.7}& {21.8}& {7.9}& {12.4}& {11.3}& {18.0}& {12.2}& {18.1}\\
         {StarCoder (15B)}& {51.7}& {29.7}& {11.4}& {21.4}& {20.2}& {29.5}& {24.5}& {26.0}\\
         {WizardCoder-SC (15B)}& {55.2}& {33.6}& {16.7}& {26.2}& {24.2}& {24.9}& {26.7}&{29.2}\\
         {CodeLlama-Python (6.7B)}& {55.3}& {34.5}& {16.4}& {19.9}& {22.3}& {17.6}& {28.5}&{28.0}\\
         {WizardCoder-CL (6.7B)}& {53.5}& {34.4}& {15.2}& {25.7}& {21.0}& {24.5}& {28.9}&{28.4}\\
         {Magicoder-CL (6.7B)}& {54.6}& {34.8}& {19.0}& {24.7}& {25.0}& {22.6}& {28.9}&{29.9}\\
         {MagicoderS-CL (6.7B)}& {\textbf{55.9}}& {40.6}& {\textbf{28.4}}& {40.4}& {28.8}& {35.8}& {37.6}&{37.5}\\
         \hline
         {WarriorCoder (6.7B)}& {55.5}&  {\textbf{41.8}}&  {26.1}& {\textbf{41.2}}& {\textbf{33.0}}& {\textbf{39.1}}& {\textbf{42.2}}& {\textbf{38.1}} \\
         \hline
    \end{tabular}
    }
    \caption{The pass@1(\%) results on the benchmark for using Python libraries in data science (DS-1000).}
    \label{tab:datascience}
\end{table*}

\begin{table*}[t!]
    \centering
    \resizebox{1.0\linewidth}{!}{
    \begin{tabular}{ccc} \hline
         {Task}& {Percentage(\%)}& {Definition}\\
         \hline
         {Code Generation}& {51.4}& {Generating source code based on certain specifications or requirements.}\\
         {Code Debugging}& {12.2}& {Identifying, diagnosing, and fixing errors or bugs in a code snippet.}\\
         {Code Optimization}& {3.8}& {Improving a program's performance, efficiency, or resource usage without changing its functionality.}\\
         {Code Reasoning}& {2.9}& {Predicting the output based on the given input or predicting the input from the known output.}\\
         {Code Analysis}& {6.6}& {Analyzing, understanding, and explaining how a piece of code works.}\\
         {Theoretical Explanation}& {22.2}& {Answering the questions about principles, theories, and properties of programming language.}\\
         {Code Transpile}& {0.9}& {Converting source code from one programming language into another programming language.}\\
         \hline
    \end{tabular}
    }
    \caption{The proportion of different tasks in the training data.}
    \label{tab:tasks}
\end{table*}

The results on the code generation benchmarks are summarized in Table~\ref{tab:overall}. \textbf{WarriorCoder} achieves SOTA performance, with a pass@1 accuracy of 80.5\% (75.6\%) in HumanEval (HumanEval +) and 76.2\% (64.8\%) in MBPP (MBPP +), surpassing all other fine-tuned models. Particularly, it shows a significant boost over on HumanEval and HumanEval+ (with gains of 32.9 and 36.0, respectively). \textbf{WarriorCoder} also outperforms MagicoderS-DS, Magicoder-DS and WaveCoder-DS, which share the same backbone architecture and similar amounts of training data. This substantial performance gap highlights the effectiveness of our approach in generating higher-quality training data, providing a clear advantage over models that rely on similar foundational setups.

Moreover, \textbf{WarriorCoder} also achieves excellent performances on the code reasoning benchmark and libraries usage benchmark. As shown in Table~\ref{tab:crux}, \textbf{WarriorCoder} outperforms a range of open-source models, including those with sizes up to 34B, in pass@1 accuracy and achieves better pass@5 accuracy compared to GPT-3.5-Turbo (66.5\% vs 63.2\% on CRUXEval-I and 66.3\% vs 59.3\% on CRUXEval-O). Table~\ref{tab:datascience} shows that \textbf{WarriorCoder} outperforms all baselines on most of the libraries, especially on SciPy, Sklearn and Tensorflow (33.0\% , 39.1\% , and 42.2\%, respectively). These results highlight \textbf{WarriorCoder} as a powerful paradigm - a data flywheel that absorbs expertise from multiple code domains. Our approach significantly enhances the target model’s ability to generalize across various tasks, demonstrating its superiority in leveraging diverse data sources to drive performance improvements.

Addiontionally, previous data flywheels typically rely on augmentations and annotations generated using proprietary LLMs and specially designed prompts. In contrast, our approach does not require pre-existing datasets, diverse handwritten prompts or proprietary LLMs. Despite this, we get competitive results that rival those of advanced proprietary code experts. This highlights the effectiveness of our data flywheel, demonstrating the feasibility of collecting high-quality data at a low cost.

\subsection{Ablation Study}
\begin{table}[t!]
    \centering
    \resizebox{1.0\linewidth}{!}{
    \begin{tabular}{ccccc} \hline
         {\#Num}& {HumanEval}&  {HumanEval+}& {MBPP}& {MBPP+}\\
         \hline
         {1}& {75.4}&  {72.6}&  {73.3}& {62.4}\\
         {2}& {77.2}&  {73.3}&   {74.5}& {62.9}\\
         {5}& {80.5}&  {75.6}&   {76.2}& {64.8}\\
         \hline
    \end{tabular}
    }
    \caption{The results observed when learning from varying numbers of experts.}
    \label{tab:ablation}
\end{table}
Table~\ref{tab:ablation} presents the results observed when the target model learns from varying numbers of experts. The target model shows a significant improvement when learning from just one code LLM, indicating that even a single code expert enables it to acquire a specific set of knowledge. However, as the number of experts increases, \textbf{WarriorCoder} benefits from learning across all expert code LLMs. As a result, the model trained with 5 code LLMs outperforms others across all four benchmarks, demonstrating the advantages of integrating knowledge from multiple specialized experts.

\subsection{Data Analysis}

\subsubsection{Dependence Analysis}
\begin{figure}[t!]
	\centering
	\includegraphics[width=1.0\linewidth]{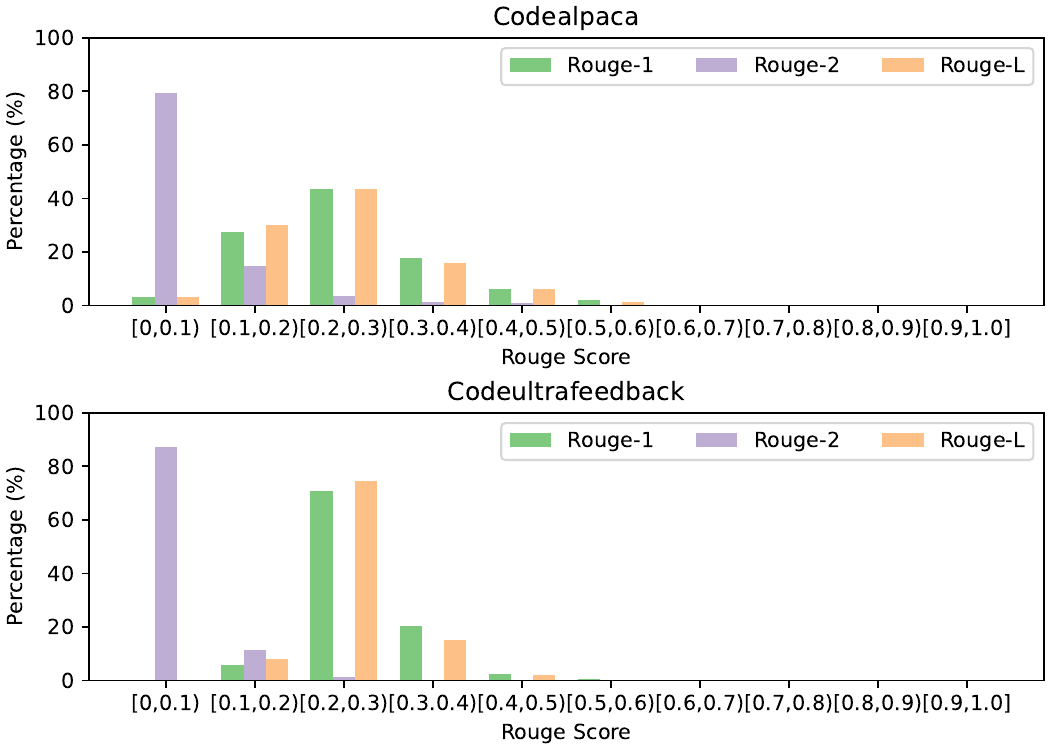}
    \caption{The overlapping rate between the mined instructions and existing training datasets.}
    \label{Fig:rouge}
\end{figure}
Figure~\ref{Fig:rouge} illustrates the overlap between the instructions mined from expert LLMs and those from widely used code training datasets, measured using the ROUGE score. The majority of the mined instructions have a ROUGE score of less than 0.3, suggesting they are largely distinct from those in existing datasets. Notably, no mined instructions exceed a ROUGE score of 0.6, further emphasizing that the mined instructions are drawn from the internal distribution of expert LLMs, rather than being simple replications or extensions of the training data. Consequently, these instructions exhibit a higher degree of independence, making them particularly valuable for training, as they provide novel examples that can enhance the target model's capabilities.

\subsubsection{Diversity Analysis}
\begin{figure}[t!]
	\centering
	\includegraphics[width=0.85\linewidth]{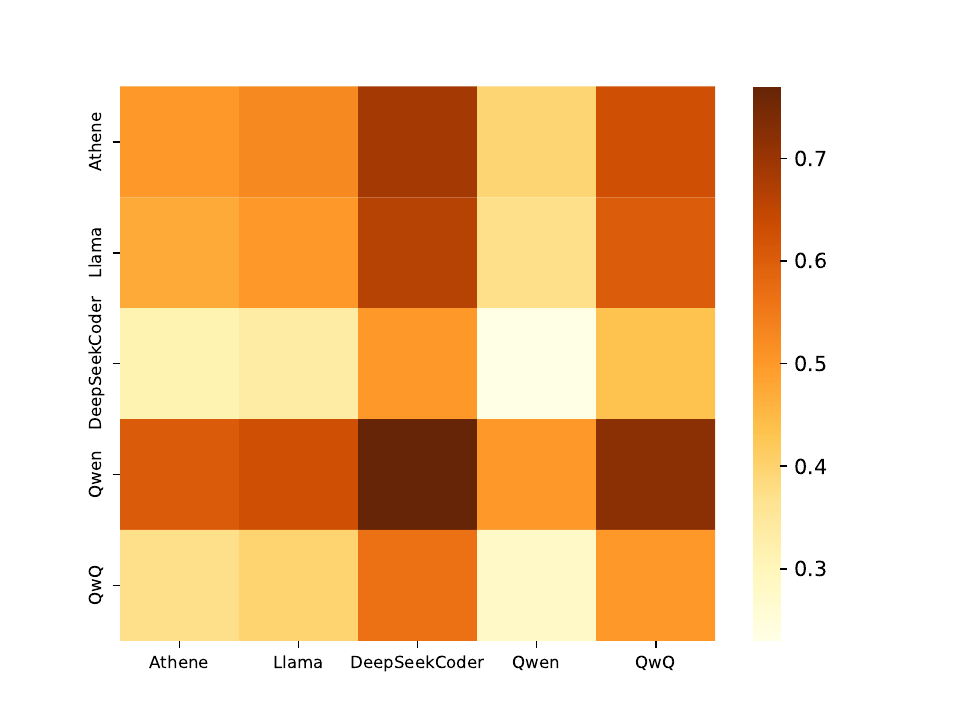}
    \caption{The heatmap of win rates of the selected code experts.}
    \label{Fig:heat}
\end{figure}
Table~\ref{tab:tasks} reveals the distribution of different tasks in the training data. The range of instructions covers a variety of tasks, ensuring that \textbf{WarriorCoder} can generalize effectively across multiple benchmarks. Notably, while Code Reasoning represents only 2.9\% of the entire dataset, \textbf{WarriorCoder} still achieves outstanding performance on CRUXEval, highlighting the potential of the framework that learns from expert battles. Furthermore, Figure~\ref{Fig:heat} illustrates the battle results between the five selected code experts. Even though an expert may have the highest Elo Rating, it is not necessarily the best performer on all tasks. However, \textbf{WarriorCoder} learns from the winner of each instruction, thereby diversifying the target responses.

\subsubsection{Difficulty Analysis}
\begin{figure}[t!]
	\centering
	\includegraphics[width=1.0\linewidth]{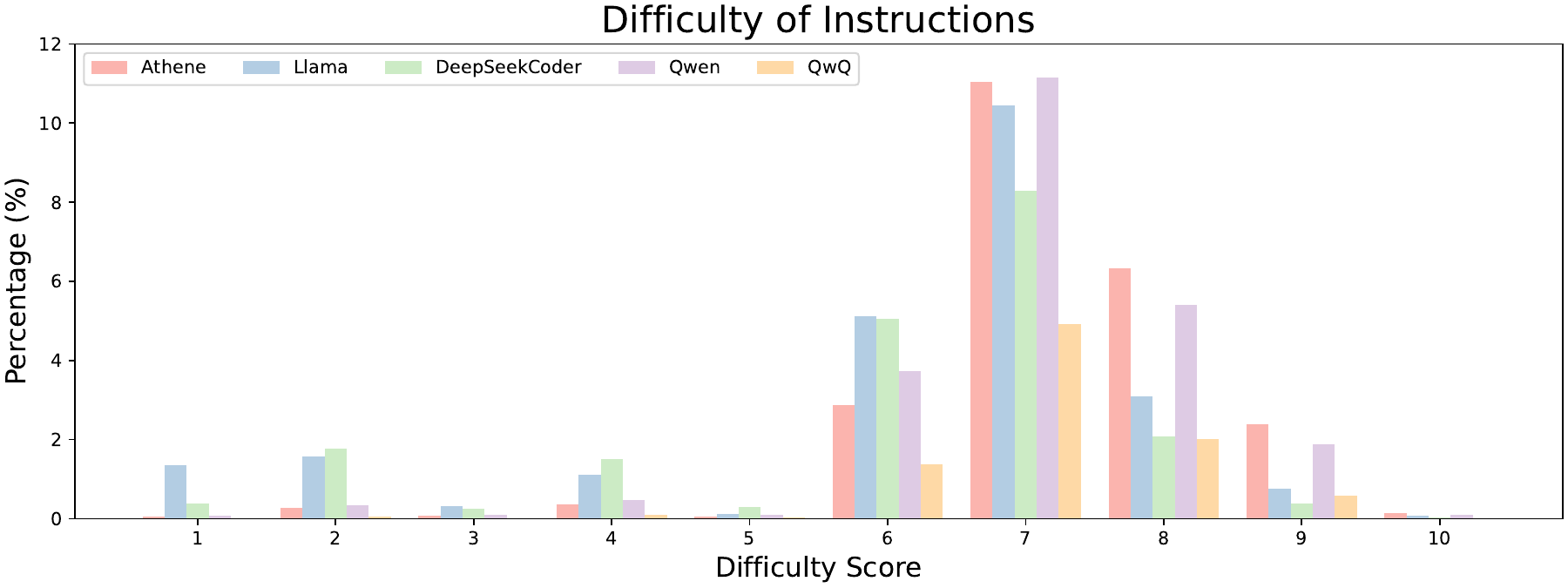}
    \caption{The proportion of difficulties of mined instructions. As mentioned in Section~\ref{sec:instruction}, the difficulties of instructions are divided into four levels: excellent (9-10), good(6-8), average(3-5) and poor(1-2).}
    \label{Fig:difficulty}
\end{figure}
Figure~\ref{Fig:difficulty} shows the difficulty distribution of the mined instructions, offering insights into the internal knowledge of the code experts. Most instructions fall within the 'good' level, with scores between 6 and 8. Instructions rated as 'excellent' (scores 9-10) constitute only a small portion of the dataset, indicating that highly complex or advanced tasks are relatively rare. Instructions with scores below 6 are excluded from the training set, as they tend to be either too easy or overly ambiguous. Such instructions are considered detrimental to the training stage, as they may not provide meaningful learning signals and could undermine the model's performance and generalization ability. More examples are listed in Appendix~\ref{sec:casestudy}.

\section{Conclusion}
This paper highlights the limitations of existing data flywheels for code LLMs that primarily rely on pre-existing datasets and annotations from a limited pool of proprietary LLMs, leading to a lack of data diversity and reinforces the systemic biases. Even more concerning is the fact that many current open-source expert code LLMs keep their training data proprietary, further restricting access to diverse and high-quality data sources. To address these challenges, we propose WarriorCoder, which learns from expert battles, enabling the absorption of each expert's strengths. Unlike existing methods that expand and refine datasets, we construct training data from scratch and achieve SOTA performances on multiple benchmarks without the need for pre-existing datasets and costly annotations. Furthermore, our approach can potentially be applied to other complex tasks besides coding in the future.

\section*{Limitations}
In this paper, we propose a novel training paradigm in which the target model learns from expert battles, aiming to overcome the limitations of current data flywheels. While we can generate high-quality and diverse data from scratch at a low cost, the battle process can become time-consuming when the number of experts is large. Exploring more efficient and effective competition modes is a promising direction for future work.

\normalem

\appendix

\section{Case Study}
\label{sec:casestudy}
Here we show two examples of mined instructions. The first has a score of 6:
\begin{center}
\fcolorbox{black}{color1}{\parbox{1.0\linewidth}{I'm trying to combine two dictionaries into one and eliminate duplicate values for a given key
the two dictionaries may have different structures, eg.\\
dictA = \{'cat1':\{'cat2':'A'\}, 'cat2':\{'cat3':'B'\}...\}\\
dictB = \{'cat1':\{'cat2':'C'\}\}\\
Combining with a recursive function seems the best option but I can't seem to get it right.\\
**Please Help.**
}
}
\end{center}
And the second has a score of 9:
\begin{center}
\fcolorbox{black}{color1}{\parbox{1.0\linewidth}{**Function to Get the Index of an Element in a 2D List (List of Lists)**\\
====================================\\
Create a function named 'get\_index\_2d' that accepts three parameters:\\
- a 2D list 'matrix'\\
- an element to search for 'target'\\
- a default value to return 'default'\\
If the 'target' exists in the 'matrix', the function should return its index (a tuple containing row, column).\\
If the 'target' is not found in the 'matrix', the function should return the 'default' value.\\
"""python\\
def get\_index\_2d(matrix, target, default=None):\\
\- \- \- \- \# implement function\\
"""\\
\#\#\# Example Use Cases:\\
|  \- \- \- \- \- \- \- \- \- \- \-Matrix \- \- \- \- \- \- \- \- \- \-| Target | Default | Expected |\\
| \- \- \- \- \- \- \-[[1,2],[3,4]] \- \- \- \- \- \-| \- \- \- \- \-4 \- \- \- \- \-| \- \- None \- \-| \- \- \- \- (1,1) \- \- \- |\\
| \- \- \- \- \- \- \-[[1,2],[3,4]] \- \- \- \- \- \-| \- \- \- \- \-5 \- \- \- \- \-| \- \- None \- \-| \- \- \- None \- \- \- |\\
| \- \- \- \- \- \- \-[[1,2],[3,4]] \- \- \- \- \- \-| \- \- \- \- \-1 \- \- \- \- \-| \- \- (0,-1) \-         | \- \- \- \- (0,0) \- \- \- |\\
| \- [['a','b'], ['c','d']] | \- \- \- 'd' \- \- | \- (-1,-1) \-  | \- \- \- \- (1,1) \- \- \- |
}
}
\end{center}

\begin{table}[t!]
    \centering
    \resizebox{1.0\linewidth}{!}{
    \begin{tabular}{p{1.0\columnwidth}}\hline
         You are an AI assistant designed to provide helpful, step-by-step guidance on coding problems. The user will ask you a wide range of Python coding questions. Your purpose is to assist users in understanding coding concepts, working through code, and arriving at the correct solutions.\\
         \hline
    \end{tabular}
    }
    \caption{The prompt for instruction mining.}
    \label{tab:promptmining}
\end{table}

\begin{table}[t!]
    \centering
    \resizebox{1.0\linewidth}{!}{
    \begin{tabular}{p{1.0\columnwidth}}\hline
         This is a chatbot arena. You will be given assistant A’s answer, and assistant B’s answer. Please act as an impartial judge and evaluate the capability of two AI assistants. You should choose the assistant that follows instructions and answers questions better. Your evaluation should consider factors such as helpfulness, relevance, and accuracy. Begin your evaluation by comparing the responses of the two assistants and provide a short explanation. Avoid any position biases and ensure that the order in which the responses were presented does not influence your decision. DO NOT allow the LENGTH of the responses to influence your evaluation, choose the one that is straight-to-the-point instead of unnecessarily verbose. When the two candidates perform equally well, choose the SHORTER answer. Do not favor certain names of the assistants. Be as objective as possible. After providing your explanation concisely within 200 words, output your final verdict by strictly following this format: "[[A]]" if assistant A is better, "[[B]]" if assistant B is better, and "[[Tie]]" for a tie. Finish your judgement within 300 words.\\
         
         [[User Question]]\\
         \{instruction\}\\
         
         [[The Start of Assistant A’s Answer]]\\
         \textcolor{red}{\{\#A’s Answer\}}\\
         
         [[The End of Assistant A’s Answer]]\\
         
         [[The Start of Assistant B’s Answer]]\\
         \textcolor{red}{\{\#B’s Answer\}}\\
         
         [[The End of Assistant B’s Answer]]\\
         \hline
    \end{tabular}
    }
    \caption{The prompt for the evaluation of pairwise competitions.}
    \label{tab:prompteval}
\end{table}

Both instructions are clear and specific, providing a solid foundation for tackling the problem. However, the first instruction is more straightforward, with fewer requirements and a greater emphasis on practical tips and strategies to handle the task. In contrast, the second instruction introduces more complex conditions and does not include any guidance, making the problem more challenging to solve.

\section{Prompts for Instruction Mining and Evaluation}
\label{sec:prompt}
Table~\ref{tab:promptmining} and Table~\ref{tab:prompteval} show the prompts for instruction mining and evaluation. To maintain impartiality in the evaluation process, we withhold the names of the opponents from the judges, thus avoiding potential system biases. Additional rules will be discussed in the next section.

\section{Rules for the Fairness of Evaluation}
\label{sec:fairness}
To ensure impartiality in the evaluation process, we establish a set of rules, including \textbf{order shuffling}, \textbf{suspicion averting}, and \textbf{offensive-defense balance}.

\textbf{Order shuffling} refers to the practice of randomizing the order in which responses from the \textbf{attacker (A)} and \textbf{defender (B)} appear in the evaluation prompt. This helps mitigate any positional bias that may arise if a particular position is favored by certain language models.

\textbf{Suspicion averting} ensures that competitors do not evaluate their own responses, preventing any potential bias in favor of their own generated answers.

\textbf{Offensive-defense balance} guarantees that all competitors have an equal number of offensive and defensive turns, maintaining fairness in the evaluation process.

\end{document}